# Improved Neural Network based Plant Diseases Identification

Ginni Garg[1] and Mantosh Biswas[2][0000-0001-9027-4432]

[1] Department of Computer Engineering, NIT Kurukshetra, India
[2] Department of Computer Engineering, NIT Kurukshetra, India
`gargginni01@gmail.com,mantoshb@gmail.com`

**Abstract.** The agriculture sector is essential for every country because it provides a basic income to a large number of people and food as well, which is a fundamental requirement to survive on this planet. We see as time passes, significant changes come in the present era, which begins with Green Revolution. Due to improper knowledge of plant diseases, farmers use fertilizers in excess, which ultimately degrade the quality of food. Earlier farmers use experts to determine the type of plant disease, which was expensive and time-consuming. In today's time, Image processing is used to recognize and catalog plant diseases using the lesion region of plant leaf, and there are different modus-operandi for plant disease scent from leaf using Neural Networks (NN), Support Vector Machine (SVM), and others. In this paper, we improving the architecture of the Neural Networking by working on ten different types of training algorithms and the proper choice of neurons in the concealed layer. Our proposed approach gives 98.30% accuracy on general plant leaf disease and 100% accuracy on specific plant leaf disease based on Bayesian regularization, automation of cluster and without over-fitting on considered plant diseases over various other implemented methods.

## 1 Introduction

As far as India is concern, the agriculture sector has a significant contribution to the GDP of India. The agriculture sector employs a large number of people all across the world. For food, a person has to depend on agriculture products. As we see in the past, to determine the type of plant diseases, farmers use the expert's people in the agriculture field, but that process is time-consuming and not always give the correct result. Because it is always true in the majority of cases that a model gives correct results as compare to humans. As we see today, work is done in the area of the Agriculture sector using Image Processing to improve the quality of products and help farmers to protect their crops from diseases, and it can extract relevant information from it. There are various methods for plant disease detection based on Artificial Neural Network (ANN) [1-4], Support Vector Machine (SVM) [5], K-Nearest

Neighbors (KNN) [6], Convolutional Neural Network (CNN) [7], Fuzzy Logic (FL) [8] and its combination. CNN based plant disease detection gives good accuracy, but it needs a large amount of data for training, which increases computational cost. SVM based plant disease detection is useful for both accurate separable and non-definite separable data but cannot return an anticipation credence value like logistic regression does. KNN based plant disease detection is useful for inputs where the probability distribution is unknown, but it is sensitive to localized data and takes considerable long computational time due to lazy learning. Fuzzy Logic-based plant disease detection is a heuristic modular way for defining any non-definite hegemony system and can attain a higher degree of cybernation but is not clearly understood and has no standard tuning and no stability criteria. Neural Network (NN) is a supervised learning approach, and it recognizes the relationship among data by a process that behaves as the human brain operates. Neural Network adjusts itself as input changes to produce the best result without the need to change the output criteria. Neural Network contains layers having inter-connected neurons or perceptrons, and the output layer has to classify input features into various classes.ANN is a non-argument model, while most emblem methods are argument models that need a higher backcloth of emblem. In NN based Bani-Ahmed et al., proposed plant disease detection method using color transformation scheme HIS and segmentation by K-mean with four clusters [1]. Other, Varthini et al., detected plant diseases using color transformation scheme HIS, threshold masking for green pixel of leaf image followed by segmentation of input image into patch size of 32*32, and classification is completing by combination of the ANN and SVM [2]. Kamlu et al., proposed the spot and catalog of plant diseases using the threshold, K-means clustering, and ANN [3]. Moreover, Pradhan et al., review the paper based on major types of neural networks with hyperspectral data [9]. So, NN based methods are resilient and can be accessible for both regression and catalog problems. It is good to miniature with no accurate data with a large number of inputs. Therefore we proposed an improved NN based plant disease identification approach, based on CIE LAB, K- means, elbow, and augmentation. The paper is sort as follows: The proposed approach for improved NN describes in Section 2 and Section 3, the results are discussing for the robustness of the proposed approach over the considered plant disease methods. Lastly, the paper cum in the conclusion section.

## 2    Proposed Method

Our proposed approach consists of various steps, firstly we do image acquisition and color transformation by RGB to CIE LAB for making images device-independent such that images captured with different color space like Standard-RGB and Adobe-RGB gives same results, secondly segmentation is used to extract lesion region using k-means and elbow, thirdly apply the augmentation in direction of 90°, 180° and 270° to increase the dataset and avoid the over-fitting, fourthly extracted the feature using 1st order and 2nd order statistics, fifthly prepared the training and testing data set and finally classify the plant diseases using improved NN. The details of the proposed steps (Fig.1.) for the identification of plant diseases are described below.

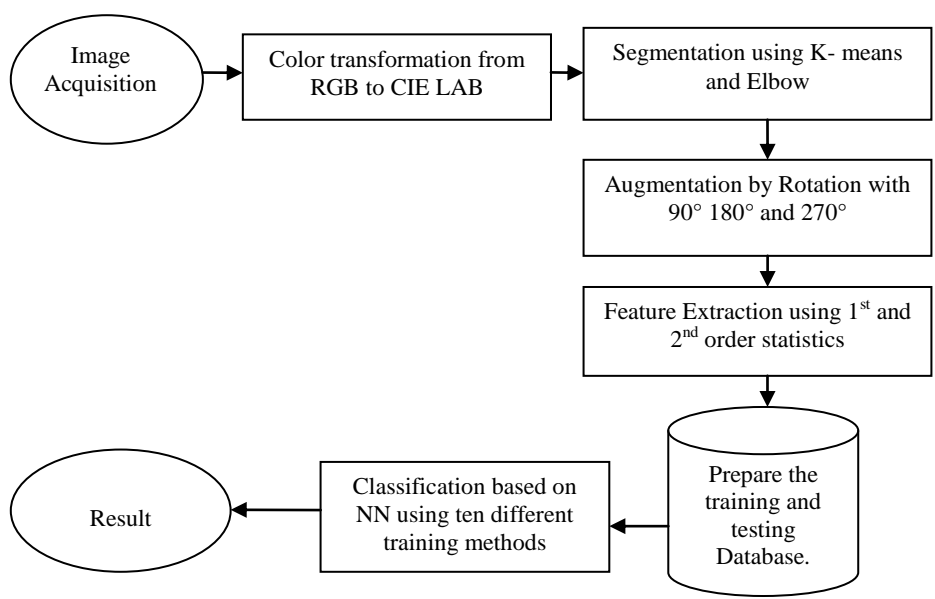

**Fig. 1.** Flow schema of the proposed modus-operandi for the identification of diseases.

### 2.1   Image Acquisition and Color Transformation

Image Acquisition is a process of collecting the required dataset to train the classifier. We have taken a total of 240 images of the widespread plant species and 516 images of the specific plant leaf species [23]. Image acquisition made through free open source like plant village, Github**,** and Internet. Color transformation means transfiguration of the description of the color from one color leeway to another [10]. We need to do color transformation from RGB to LAB because RGB is device-dependent whereas L*a*b color system is device independent, so that on transforming from s-RGB or Adobe-RGB to LAB produces the same set of 3-coordinates, so researcher working with s-RGB or Adobe-RGB compute same accuracy i.e., independent of device used to capture the image of plant leaf. Moreover, the objective of the color leeway is to proficiency the enumeration of colors in some caliber way. CIE L*a*b color model perceives colors like human beings. This step is relevant in the feature extraction process because s-RGB and Adobe-RGB color-space give different pixel values for an image, whereas LAB color space gives the same pixel values. Some of the disease leaf test images in s-RGB, as shown in Fig.2.

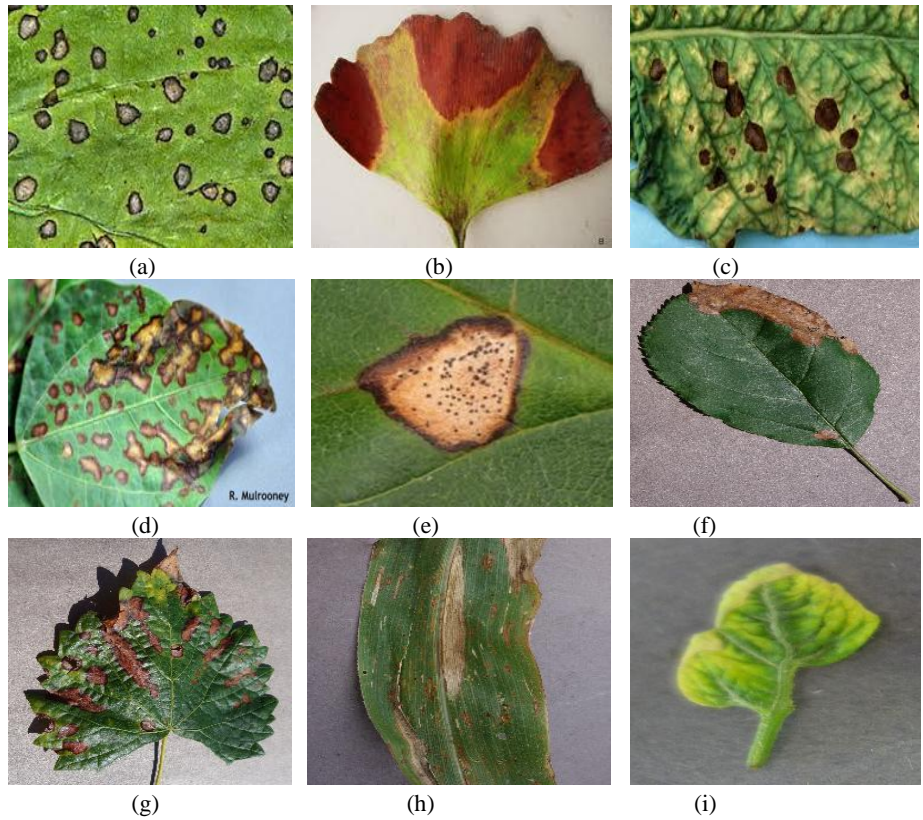

**Fig. 2.** Disease leaf (a) Frog leaf spot (b) Early Scorch (c) Fungal Disease (d) Bacterial Disease and (e) Sunshine Burn (f) Apple Rot (g) Grape ESCA (h) Maize Blight and (i) Tomato Yellow_Virus

### 2.2  Segmentation

Image septation in the catalog of an image into various categories. Proposed approach to identify effected areas under study. Many scientists have worked in the area of image segmentation using clustering, and one of modus operandi is K-means clustering, which is used in our proposed approach aims at minimizing an objective function [1, 3]. Moreover, we used Elbow for the automation of the clusters [11]. In Fig.3. represents the clustering of the bacterial plant leaf using the K-means with Elbow, in which K-means produced 6-clusters for the input image.

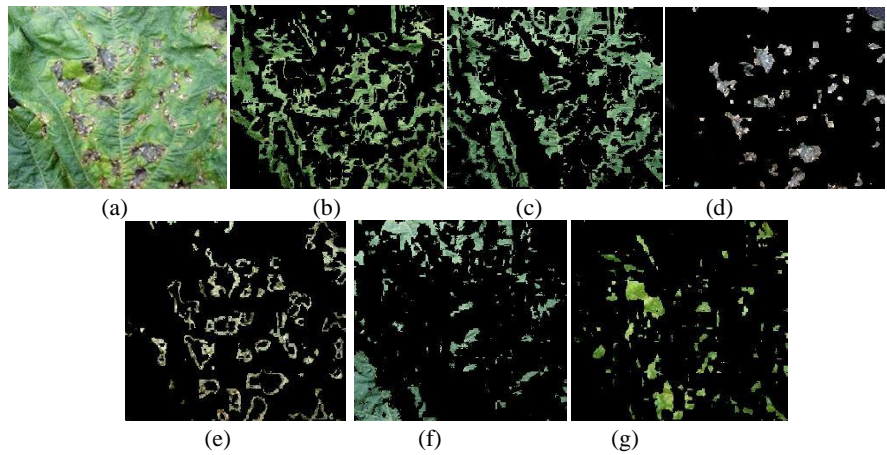

**Fig. 3.** Bacterial Leaf Disease (a) Original (b) Clutch-1 (c) Clutch-2 (d) Clutch-3 (e) Clutch-4 (f) Clutch-5 (g) Clutch-6

### 2.3 Augmentation

Augmentation is an approach used to increase the dataset and overcoming the problem of over-fitting [12]. This method is used for functional prediction of images when images are given withdifferent orientation then orientation with which classifier is trained. In our proposed approach, we use rotated images by 90°, 180°, and 270° (Fig.4. and Fig.5.).

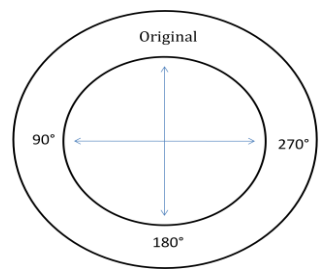

**Fig. 4.** Augmentation approach by rotating 90°, 180°, 270°.

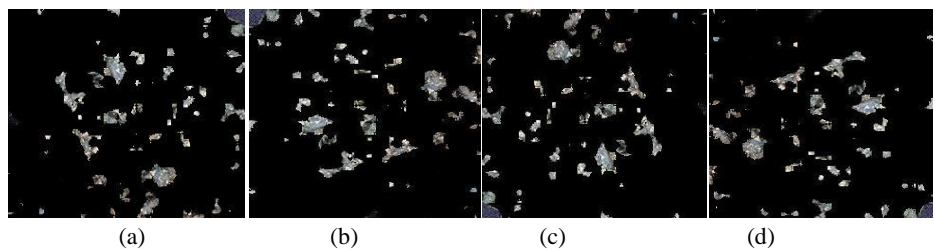

**Fig. 5.** Rotated lesion image (a) Original (b) 90° (c) 180° and (d) 270°.

## 2.4 Feature Extraction

Features are the measurable descriptors of an object, and it describes elementary characteristics such as shape, color, texture, or motion, among others, and use for classifying the type of plant diseases [13]. In our proposed work, we use $1^{st}$ order emblem to approximate properties of individual pixel values and $2^{nd}$ order emblem for the spatial relationship between image pixels. We use $1^{st}$ order statistics for Mean, Kurtosis, Standard Deviation, Variance and Skewness, and $2^{nd}$ order statistics for contrast, correlation, energy, homogeneity, smoothness, IDM, RMS, and entropy (Table 1.).

**Table 1.** Represents the thirteen different features.

| Feature Name | Formula |
| --- | --- |
| Contrast | $C = \sum_{x,y=1}^{A,B} |x-y|^2 q(x,y)$ ; Here, q(x,y) GLCM, x & y are row & column and A is total rows and B is total columns. |
| Correlation | $Corr = \sum_{x,y=1}^{A,B} \big((x-\mu)(y-\mu)q(x,y)\big)/(\sigma(x) * \sigma(y))$ ; Here, mean is $\mu$ and standard deviation is $\sigma$. |
| Energy | $E = \sum_{x,y=1}^{A,B} q(x,y)^2$ |
| Homogeneity | $H = \sum_{x,y=1}^{A,B} q(x,y)/(1+|x-y|)$ |
| Mean | $\mu = \frac{1}{AB} * \sum_{x=1}^{A}\sum_{y=1}^{B} q(x,y)$ |
| Standard Deviation | $\sigma = \sqrt[2]{\frac{1}{AB} * \sum_{x=1}^{A}\sum_{y=1}^{B} (q(x,y)-\mu)}$ |
| Kurtosis | $K = \{\frac{1}{AB} * \sum_{x=1}^{A}\sum_{y=1}^{B} ((q(x,y)-\mu)/\sigma)\textasciicircum 4\} - 3$ |
| Skewness | $S = \frac{1}{AB} * \sum_{x=1}^{A}\sum_{y=1}^{B} ((q(x,y)-\mu)/\sigma)$ |
| Variance | $Var = \frac{1}{AB} * \sum_{x=1}^{A}\sum_{y=1}^{B} (q(x,y)-\mu)$ |
| Smoothness | $R = 1 - 1/(1+\sigma^2)$ |
| IDM | $HH = \sum_{x,y=1}^{A,B} q(x,y)/(1+|x-y|)$ |
| RMS | $y = \sqrt[2]{\sum_{x,y=1}^{A,B} (|q(x,y)|)^2/A}$ |
| Entropy | $h = -\sum_{x,y=1}^{A,B} q(x,y)(\log q(x,y))$ |

## 2.5 Training and Testing Set

The training set is used to fit the parameters of the miniature, whereas the testing set is used to deliver a fair estimation of a final miniature fit on the training dataset. In our proposed set includes a database of features and output classes used to train and test the classifier. We use in improved NN of 13-Features, and 5-Output classes, representation as V and OC, respectively describe below:

$$V = F1, F2, F3, \ldots\ldots F13 \tag{1}$$

$$OC = C1, C2, C3, \ldots\ldots C5 \tag{2}$$

## 2.6 Classifier

Classification means to identify the type of category a new observation belongs to the set of known categories. In our proposed approach, we are used NN as classifier to identify the disease from leaf image and use various training algorithms in our improved NN is given below:

**Bayesian Regularization.** Bayesian regularization is a mathematical procedure that converts a nondefinite regression into a "well-posed" emblem problem in the way of a chine regression [12].

**Levenberg-Marquard.** The LM fitting of a curve procedure is an amalgam of two diminution procedure: the Gauss-Newton procedure and the gradient descent procedure [14].

**BFGS Quasi-Newton.** Quasi-Newton method is used to find the local minima or maxima as a substitute to the Newton method. Usually, it is used if the Jacobian or Hessian matrix is difficult to compute or is unavailable [15].

**Resilient Back propagation.** Resilient back propagation is a supervised learning method used in the feed-forward neural network [16].

**Scaled Conjugate Gradient**. It is a training function used for the solution of the set of linear equations whose matrix is symmetric and positive definite [17].

**Powell/Beale Restarts Conjugate Gradient.** The Beale-Powell restart method is used for large-scale depraved boost [18].

**Conjugate Gradient with Fletcher-Powell.** The Davidon-Fletcher-Powell penalty function method is an approach that has been used successfully to solve constrained minimization problems. The method devised by combining an exterior penalty function with a performance function for solving constrained minimization problems [19].

**Polak-Ribiere Conjugate Gradient.** It is a method based on the conjugate gradient used as a training function in the neural network [20].

**One Step Secant.** This method lies in between conjugate gradient descent and the quasi-newton method. This algorithm usually need less computation as compare to the BFGS algorithm [21].

**Variable Learning Rate Back propagation.** In reality, learning rate plays a significant role in the training algorithm such that proper choice of the learning rate is essential to have good accuracy such that the above method vary learning rate until one can get good accurate results [22].

## 3     Results

Simulation of our proposed approach and existing plant disease detection on [1], [2] and [3] performed on five general plant disease, namely, Bacterial leaf spot, Early scorch, Frog-Leaf spot, Fungal disease, and Sunburn disease (Fig.1.) and experiment is also extended on the specific plant diseases which include Grape, Tomato, Apple and Maize (Fig.2. and Table 5.). Performance evaluation is done by Accuracy (Table 4.& 7.), which is given by the below equation (3), where TP is True Positive, TN is True Negative, FP is False Positive and FN is False Negative.

$$Accuracy = (TP+TN)/(TP+TN+FP+FN) \qquad (3)$$

In our approach, general plant species and specific plant leaf species taken in ratio 7:3 for training & testing (Table 2.& 5.). In improved NN, we have given the input that consists of 13 features, Epoch is 1000, performance function is Mean Squared Error (MSE), 5 output classes (general diseases), and activation function is sigmoidal in concealed layer and precarious in output layer. All the plant disease experimental computation is done in R2017b MATLAB toolbox with a personal computer of 4 GB memory, Window 10 64-bit operating system, and Intel (R) Core (TM) i3-6006U CPU @ 2.00 GHz.

Tables 2 and 5 are represented the general and specific, respectively, the dataset of the considered plant leaf for this experiment, which shown with the number of training and testing images.

**Table 2.** Represents the thirteen different features.

| Category | Training | Testing |
|---|---|---|
| Bacterial Leaf Spot | 36 | 16 |
| Early Scorch | 39 | 17 |
| Frog Eye Spot | 42 | 18 |
| Fungal | 36 | 16 |
| Sunburn | 14 | 6 |

Table 3 represents the computation used ten-different training algorithms to train NN and different combinations of Neurons used in the only concealed layer i.e., 5-10-15-20. From this table, we found that Bayesian Regularization gives excellent results over other training algorithms. All percentage compute by taking an average of 10 fold observations of confusion matrix.

**Table 3.** Represents the accuracy of thirteen different algorithms with neurons N: 5, 10, 15, 20.

| Training Algorithm Name | No of Neurons in only hidden layer (N) | | | |
|---|---|---|---|---|
| | N=5 | N=10 | N=15 | N=20 |
| Bayesian Regularization | **93.17** | **96.88** | **97.38** | **98.30** |
| Levenberg-Marquardt | 86.35 | 93.02 | 96.18 | 95.56 |
| BFGS Quasi-Newton | 47.85 | 55.83 | 64.31 | 53.97 |
| Resilient Backpropagation | 54.42 | 74.80 | 72.25 | 79.39 |
| Scaled Conjugate Gradient | 53.26 | 73.47 | 74.04 | 89.53 |
| Conjugate Gradient with Powell/Beale Restarts | 55.92 | 70.26 | 73.03 | 85.06 |
| Fletcher-Powell Conjugate Gradient | 53.38 | 64.39 | 67.37 | 76.76 |
| Polak-Ribiére Conjugate Gradient | 55.55 | 69.83 | 79.55 | 89.08 |
| One Step Secant | 55.21 | 60.05 | 61.29 | 68.44 |
| Variable Learning Rate Backpropagation | 57.56 | 66.71 | 69.65 | 66.80 |

Table 4 represents the accuracy of the considered general plant disease detection method [1], [2] and [3] and the proposed approach. From this table, it describes that the proposed improved NN gives the best accuracy results over [1], [2], and [3].

**Table 4.** Comparison of results on considered dataset with proposed method and other results.

| [1] | [2] | [3] | **Proposed with Bayesian Regularization (N=20)** |
|---|---|---|---|
| 93.05 | 87.52 | 88.67 | **98.30** |

Table 5 Represent the dataset of the specific plant leaf for this experiment, which shown with the number of training and testing images.

**Table 5.** Comparison results on considered dataset with proposed method and other results.

| **Plant Species** | **Diseases Category** | **Images Training** | **Images Testing** |
|---|---|---|---|
| Grape | Black_rot | 5 | 3 |
|  | Black_Measles (Esca) | 34 | 14 |
| Tomato | Bacterial Spot | 17 | 7 |
|  | Early_blight | 31 | 13 |
|  | Late_blight | 25 | 11 |
|  | Septoria_spot | 11 | 5 |
|  | Yellow_Curl_Virus | 17 | 7 |
| Apple | Cedar_Rust | 81 | 35 |
|  | Black_Rot | 76 | 32 |
| Maize | Nothern_Blight | 39 | 17 |
|  | Common_Rust | 25 | 11 |

Table 6 represents the results of Accuracy for proposed improved NN using Bayesian Regularization on the dataset shown in Table 5.

**Table 6.** Comparison results on considered dataset with specific plant diseases.

| **Input Feature** | **Training Algorithm** | **Performance Function** | **Epoch** | **Neurons** | **Output** | **Accuracy** |
|---|---|---|---|---|---|---|
| 13 | Bayesian Regularization | Mean Square Error | 1000 | 1 | 2 | 100 |
|  |  |  |  | 15 | 5 | 100 |
|  |  |  |  | 1 | 2 | 100 |
|  |  |  |  | 1 | 2 | 100 |
|  |  |  |  |  | **Total** | **100%** |

Table 7 represents the accuracy of the species considered plant disease detection method [1], [3], and the proposed approach. From this table, it concludes that the proposed improved NN gives the best accuracy results over [1] and [3].

411

Table 7. Results of Accuracy for specific plant disease identification.

| [1] | [3] | **Proposed with Bayesian Regularization** |
|---|---|---|
| 90.58 | 72.33 | **100.00** |

## 4      Conclusion

In this paper, we proposed an improved NN based plant disease identification. Our proposed approach includes automation for the clusters, proper identification of the lesion region, augmentation, and best training algorithm to train the neural network. Experimental results show that the choice of the training algorithm is vital to have good results of accuracy, and augmentation helps in increasing the dataset and overcoming the over-fitting problem. The overall proposed approach gives excellent accuracy of 98.30% on testing the widespread plant diseases and 100.0% on testing the specific plant diseases over a considered dataset. Further accuracy can be improved for disease detection in plant leaf with the help of the fusion of various classifiers and integration of the fuzzy logic to the neural network.